\documentclass{bmvc2k}

\usepackage{amsfonts} 
\usepackage{caption}
\usepackage{booktabs}
\usepackage{array}
\usepackage{float}
\usepackage{arydshln}

\title{Conditional Human Sketch Synthesis with Explicit Abstraction Control}

\addauthor{Dar-Yen Chen}{chendaryen@outlook.com}{}
\addauthor{Yi-Zhe Song}{y.song@surrey.ac.uk}{1}
\addinstitution{
 SketchX, CVSSP\\
 University of Surrey, UK
}

\runninghead{Chen, Song}{Human Sketch Synthesis with Abstraction Control}

\def\etal{\emph{et al}\bmvaOneDot}

\begin{document}

\maketitle

\begin{figure}[h]
\begin{center}
    \includegraphics[width=1.\linewidth]{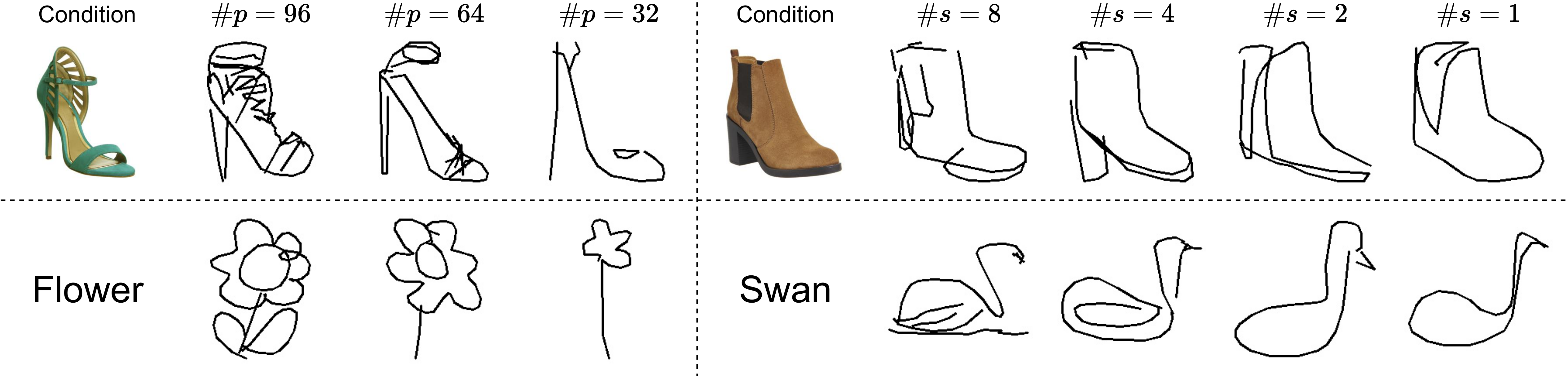}
\end{center}
   \caption{Our approach enables explicit control over the length of points (left) and strokes (right) in both photo-to-sketch synthesis (top) and class-conditional synthesis (bottom). This allows us to create recognizable sketches with few points (4th column) or a single stroke (9th column). Our results maintain key features at various abstraction levels, adjusting the detail amount accordingly. $\#p$, $\#s$: number of points and strokes in the sketch, respectively.}
\label{fig:banner}
\end{figure}

\begin{abstract}
This paper presents a novel free-hand sketch synthesis approach addressing explicit abstraction control in class-conditional and photo-to-sketch synthesis. Abstraction is a vital aspect of sketches, as it defines the fundamental distinction between a sketch and an image. Previous works relied on implicit control to achieve different levels of abstraction, leading to inaccurate control and synthesized sketches deviating from human sketches. To resolve this challenge, we propose two novel abstraction control mechanisms, state embeddings and the stroke token, integrated into a transformer-based latent diffusion model (LDM). These mechanisms explicitly provide the required amount of points or strokes to the model, enabling accurate point-level and stroke-level control in synthesized sketches while preserving recognizability. Outperforming state-of-the-art approaches, our method effectively generates diverse, non-rigid and human-like sketches. The proposed approach enables coherent sketch synthesis and excels in representing human habits with desired abstraction levels, highlighting the potential of sketch synthesis for real-world applications.
\end{abstract}

\section{Introduction}
\label{sec:intro}
From prehistoric cave paintings to modern digital art, sketching has been an innate human ability, encapsulating our experiences and emotions in simple strokes. Sketching encompasses two fundamental aspects: content and style. Content refers to the subject matter of the sketch. Abstraction, ranging from a single stroke to a detailed depiction, lies at the core of the style. This study aims to explore whether neural networks can emulate human sketching capabilities by conditionally synthesizing sketches with a desired level of abstraction.

The field of sketch synthesis has made significant strides since the introduction of Sketch-RNN \cite{sketch_rnn}, which marked the beginning of the pursuit to generate sketches in a human-like manner. Despite the progress made in the field, previous works predominantly focus on unconditional synthesis within a single category or a limited number of categories. There has been limited research on class-conditional sketch synthesis models. Photo-to-sketch synthesis, a relatively unexplored area, has witnessed only a few attempts by researchers, including works by Song \etal \cite{Song_2018_CVPR}, Muhammad \etal \cite{Muhammad_2018_CVPR}, SketchLattice \cite{sketchlattice}, and CLIPasso \cite{clipasso}. Although some of these works \cite{Muhammad_2018_CVPR, sketchlattice, clipasso} propose different abstraction controlling mechanisms, they suffered from drawbacks such as incomplete sketches, inaccurate abstraction control, and rigid styles.

The limitations of past works come from the inability to learn from data how humans sketch at different abstraction levels. They need to rely on implicit support like lattice representation \cite{sketchlattice} or pretrained classifiers \cite{Muhammad_2018_CVPR, clipasso}. Conversely, our explicit control approach offers an accurate and human-like solution for controlling the abstraction level in synthesized sketches. We propose two abstraction control mechanisms, state embeddings and stroke token, which integrate with transformer-based \cite{transformer} latent diffusion models (LDM) \cite{dit}. These mechanisms enable accurate control over the length of points and strokes, respectively. By feeding the number of points or strokes to the diffusion model straightforwardly, our approach can adapt the sketch strategy based on explicit abstraction level constraints, generating sketches with diverse styles that emulate human traits (see Figures \ref{fig:compare} and \ref{fig:compare_stroke}). To reduce the computational burden on the diffusion backbone, we pretrain a Variational Autoencoder (VAE) \cite{vae} to provide compressed neural representations of sketches, significantly shortening sketches and allowing more efficient leveraging of the transformer. By incorporating adaptive layer normalization and cross-attention conditioning in diffusion models, we seamlessly achieve class-conditional and photo-to-sketch synthesis.

In summary, our contributions include: (i) proposing state embeddings and the stroke token to accurately and naturally control the number of points and strokes in the generated sketches, respectively. (ii) presenting a transformer-based LDM framework that accomplishes class-conditional synthesis and photo-to-sketch synthesis. (iii) being the first model that succeeds in class-conditional sketch synthesis across a wide range of categories. (iv) enhancing the real-world applicability of sketch synthesis by providing controllability of both content and abstraction level.

\section{Related Work}

\textbf{Free-Hand Sketch Synthesis}
Sketch-RNN \cite{sketch_rnn} initiated the study of neural representations of vector sketches by using a sequence-to-sequence Variational Autoencoder (seq2seq-VAE) \cite{seq2sqq_vae} combined with a recurrent neural network-based autoregressive model. Addressing the issue of long sketch point sequences, Das \etal \cite{bazier_sketch} employed Bézier curves to represent strokes, allowing for fewer points to control smooth strokes compared to point-by-point segments. This approach leads to improved performance in synthesizing longer sketches.

Recent advancements introduced Denoising Diffusion Probabilistic Models (DDPMs) \cite{ddpm} for image synthesis, showcasing their potential in generating high-quality images. SketchKnitter \cite{sketchknitter} extended DDPMs to free-hand sketch synthesis, demonstrating the framework's benefits in the sketch domain. However, a significant limitation of these studies is their models' need for single-category training or their ability to perform only unconditional synthesis on multi-category datasets. Class-conditional sketch synthesis remains unexplored in the literature.

For instance-based synthesis, Song \etal \cite{Song_2018_CVPR} employed deep neural networks with shortcut cycle consistent constraints to achieve photo-to-sketch synthesis. Muhammad \etal \cite{Muhammad_2018_CVPR} trained a reinforcement learning agent to automatically remove strokes from edge maps. Qi \etal \cite{sketchlattice} proposed a lattice representation of sketches, combining Long Short-Term Memory (LSTM) \cite{lstm} with graph models to create SketchLattice, capable of generating sketches based on edge maps. Taking a different approach, CLIPasso \cite{clipasso} implemented a step-by-step optimization algorithm for sketch synthesis.

\textbf{Sketches Abstraction}
Muhammad \etal \cite{Muhammad_2018_CVPR} developed a reinforcement model that employs a pretrained classifier to provide recognizability as a reward, allowing the model to remove unimportant strokes and achieve various abstraction levels. SketchLattice \cite{sketchlattice} adjusts the abstraction level by controlling the density of lattice points. Another approach, CLIPasso \cite{clipasso}, sets the number of strokes during initialization, with the optimization process guided by CLIP \cite{clip} similarity, influencing stroke shapes. However, a common issue among these methods is that the resulting sketches do not reflect human habits. It is essential to consider that people alter their sketching styles when applying different abstraction levels, a factor that should be incorporated into future models. Neither classifier, lattice, nor CLIP can express this aspect of human sketching behavior.

\textbf{Diffusion Probabilistic Models}
In recent years, DDPMs \cite{ddpm, diffusion_beat_gan, ddpm_improved} have gained prominence as an effective approach to image synthesis. DDPMs treat image generation as a gradual Markov denoising process, facilitating the capture of major data variations. Nonetheless, the time-consuming nature of the multi-step denoising process remains a challenge. DDIM \cite{ddim} addressed this issue by proposing a non-Markov process for faster sampling and inference. Further advancements in DDPM efficiency were made by Rombach \etal \cite{ldm} through the integration of VAEs. By capitalizing on VAEs' dimensional reduction capabilities, the time consumption of the denoising process during training and synthesis can be significantly reduced. Although DDPMs have been applied to various sequential data types, such as speech \cite{diffwave}, their use in sketch synthesis is relatively unexplored, with only Diff-HW \cite{diff_hw} and SketchKnitter \cite{sketchknitter} venturing into this domain, indicating that the field remains in its infancy.

\section{Method}

In our method, we represent a sketch as a sequence of $N$ points, denoted as $S = (s_1, s_2, …, s_N)$. Each point $s_i = (x_i, y_i, p_i)$ is defined by its normalized absolute position, i.e., $x_i, y_i \in [-1, 1]$, and the corresponding binary pen-state $p_i$. The pen-state is "draw" when $p_i = 0$, and "pen lift" when $p_i = 1$. To maintain a consistent shape for all sketches, we keep $N$ constant across the dataset. We apply the Ramer–Douglas–Peucker algorithm \cite{ALGORITHMSFT} to simplify strokes. For shorter sketches, we pad them to extend their length, while for longer strokes, we truncate the sequence to the first $N$ points.

\subsection{Latent Diffusion Framework}

In this study, we employ transformer-based Latent Diffusion Models (LDM) as our synthesis framework. Contrary to traditional Diffusion Models, as used in Diff-HW \cite{diff_hw} and SketchKnitter \cite{sketchknitter}, which operate directly on sketch points, our method compresses the temporal dimension while maintaining the semantics of the sketch through a VAE, thus creating a compact low-dimensional latent space. The transformer \cite{transformer} architecture naturally fits the sequential structure of sketch vectors.

By focusing on the VAE's latent space, the diffusion backbone effectively reduces the processing of redundant information in high-dimensional spaces, which in turn significantly decreases time and memory consumption during training and inference. Let the encoder and decoder of the first-stage VAE be represented by $E$ and $D$, and the diffusion backbone by $\epsilon_\theta$. LDMs are perceived as sequential denoising autoencoders $\epsilon_\theta(z_t,t), t = 1, ..., T$, trained to estimate the noise component in $z_t$ and generate a less noisy $z_{t-1}$. $z_t$ is acquired via a diffusion process applied to $z_0 = E(S)$. This process is characterized as a Markov Chain of length $T$, with each step involving a minor Gaussian perturbation of the prior state. We will use $\epsilon_\theta(z_t)$ as an abbreviation for the time-dependent $\epsilon_\theta(z_t,t)$.

For conditional synthesis, $\epsilon_\theta(z_t)$ can be reformulated as $\epsilon_\theta(z_t,c)$, given the condition $c$. With the aid of a reweighted variational lower bound \cite{diffusion_beat_gan}, the objective can be formulated as:

\begin{equation}
\mathcal{L} = \mathbb{E}_{z, \epsilon \sim \mathcal{N}(0,\mathbf{I}), t, c}\left[\lVert \epsilon - \epsilon_\theta(z_t, c) \rVert_2^2\right]
\label{eq:loss_c_ldm}
\end{equation}

where $t \sim \mathcal{U}({1, ..., T})$. The likelihood learning is achieved by minimizing the mean-squared error between the actual noise $\epsilon$ and the estimated noise $\epsilon_\theta(z_t, c)$.

\begin{figure}[t!]
\begin{center}
    \includegraphics[width=1.\linewidth]{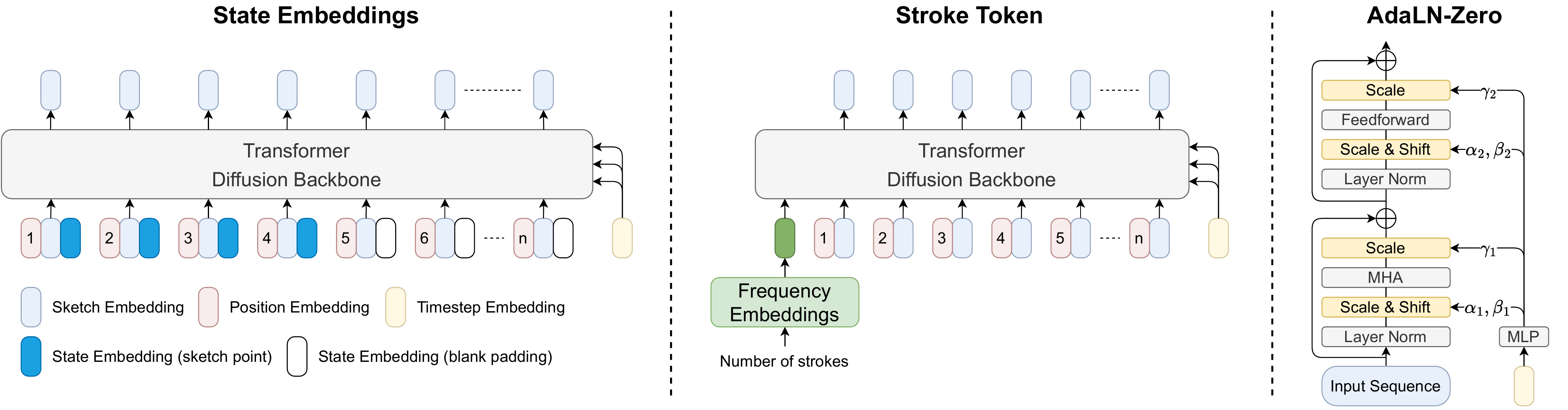}
\end{center}
   \caption{Conditioning mechanism architectures. Left: Our state embeddings represent the state of each sketch token in the input sequence, conveying point length information to the model. Middle: We utilize stroke embedding as an extra token to the input sequence, allowing it to interact with the sketch sequence via MHA. This interaction is critical for effective sketch synthesis with varying stroke counts. Right: Details of adaLN-Zero \cite{dit}.}
\label{fig:mechanisms}
\end{figure}

We use AdaLN-Zero, proposed by Peebles and Xie \cite{dit}, for processing timestep embeddings. AdaLN-Zero has proven to be more effective than conventional adaptive layer normalization in transformer blocks for diffusion models, as it regresses dimension-wise scaling parameters $\gamma$ prior to the residual connection, in addition to regressing scale $\alpha$ and shift $\beta$ parameters after layer normalization (see Figure \ref{fig:mechanisms}).

\subsection{Conditional Synthesis}

\textbf{Class-conditional synthesis}
By summing label embeddings with timestep embedding, AdaLN-Zero allows us to efficiently inject category information into the diffusion backbone, achieving class-conditional sketch synthesis.

\textbf{Photo-to-Sketch synthesis}
We leverage the pretrained CLIP (Contrastive Language-Image Pretraining) \cite{clip} image encoder as the photo encoder. CLIP is trained using contrastive learning on a wide variety of image domains and linguistic concepts, effectively capturing the semantics in images. We concatenate the classification tokens from all self-attention layers in CLIP ViT-B/32 to construct a length-12 sequence. To exploit CLIP's capabilities, we attach multi-head cross-attention layers following each multi-head self-attention (MHA) layer in the transformer architecture, which is similar to the conditioning used in Stable Diffusion \cite{ldm} for text-to-image tasks. These cross-attention layers enable our model to capture the object's semantics and establish stronger correspondence between the photo and sketch domains, resulting in coherent synthesized sketches that resemble the reference photo.

\subsection{Controlling Abstraction}

In sketches, the level of abstraction can be determined by two factors: point length and stroke length. Point length refers to the total number of points in a sketch, which directly impacts its complexity and abstraction. Conversely, stroke length pertains to the number of strokes that make up a sketch. While there is no definitive relationship between stroke length and point length, they generally exhibit a positive correlation. For humans, manipulating stroke length is more intuitive than adjusting point length, as strokes naturally form during sketching, whereas points are a byproduct of computer graphics. Prior research by Song \etal \cite{Song_2018_CVPR} and CLIPasso \cite{clipasso} demonstrated precise control over stroke count in generated samples, but no existing approach has managed to regulate sketch point length.

In this work, we accurately control both point length and stroke length, enabling more refined manipulation of sketch abstraction levels. The proposed state embeddings and stroke token architectures are illustrated in Figure \ref{fig:mechanisms}.

\textbf{Point length}
We introduce learnable state embeddings inspired by the segmentation embeddings used in BERT \cite{bert}. There are two types of state embeddings: one representing sketch points and another representing blank padding. These state embeddings, combined with the input sequence and position embeddings, enable the model to differentiate sketch points from padding.

\textbf{Stroke length}
In order to achieve stroke length control in the sketch generation model, we first encode the desired number of strokes using frequency embeddings. The resulting stroke embedding is then concatenated to the input sequence as an additional token, referred to as the stroke token. By incorporating the stroke token, it can interact with the sketch sequence through multi-head attention within the transformer layers. This allows the model to adjust the sketch sequence based on the information encoded in the stroke token, effectively controlling the stroke length in the generated sketches.

\section{Experiments}
In this section, we provide evaluations in Section \ref{eval}, comparison with SOTA \cite{Muhammad_2018_CVPR, sketchlattice, clipasso} in Sections \ref{compare_qual} and \ref{compare_quan}, and further ablative experiments in Section \ref{ablation}.

\textbf{Dataset}
We evaluate our approach on two datasets: the \emph{Quick, Draw!} dataset \cite{sketch_rnn} and the QMUL Shoe-V2 dataset \cite{sketch_shoe}. The \emph{Quick, Draw!} dataset is a large-scale collection of hand-drawn sketches, comprising 345 categories with over 50M sketch vectors. For our experiments, we selected 75 categories to train the first-stage VAE and class-conditional models. Each category consists of 50K training data and 500 testing data, resulting in a total of 3.75M training data and 37.5K testing data. The QMUL Shoe-V2 dataset contains photo and sketch vector pairs, with 5,982 training data samples and 666 testing data samples. We use the QMUL Shoe-V2 dataset to train and evaluate our photo-to-sketch models.

\textbf{Implementation Details}
In this work, all models are trained using a single Nvidia 3080 Ti GPU.  We set the number of points, $N$, to 192. We employ a first-stage VAE with a compression ratio of $4\times$, resulting in a sketch latent representation of length 48 for processing by the diffusion backbone. As the QMUL Shoe-V2 dataset is relatively small, we initialize the overlapping weights of our photo-to-sketch models using class-conditional models and apply Low Rank Adaptation (LORA) \cite{lora} to mitigate overfitting. Lastly, all results in this section are sampled with the full 1000 steps.

\begin{figure}[t]
\begin{center}
    \includegraphics[width=1.\linewidth]{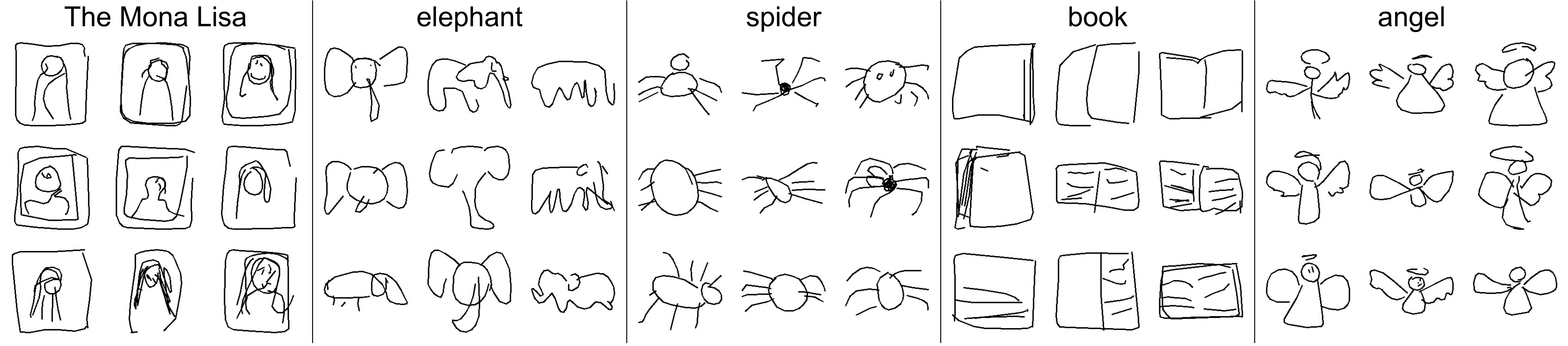}
\end{center}
   \caption{Class-conditional synthesis results showcasing diverse sketches across multiple categories.}
\label{fig:class}
\end{figure}

\begin{figure}[t]
\begin{center}
    \includegraphics[width=.8\linewidth]{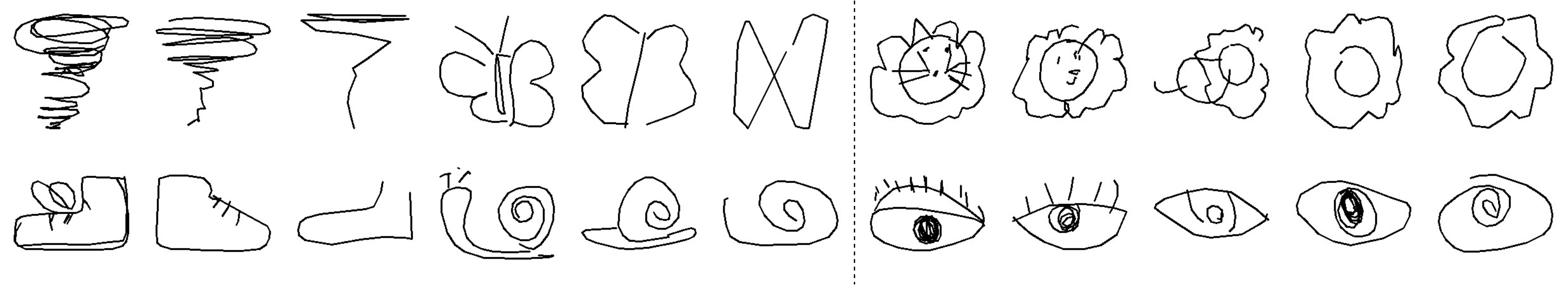}
\end{center}
\caption{Abstraction level controlling on class-conditional synthesis. Left: Point length controlling. Right: Stroke length controlling.}
\label{fig:class_abstract}
\end{figure}

\begin{table}[t!]
\begin{center}
    \begin{tabular}{lcccc}
        \toprule
          & Class/Point & Class/Stroke & Photo/Point & Photo/Stroke \\
        \midrule
        Acc. (\%)         & 99.3 & 96.9  & 99.8 & 90.6 \\
        Acc. $\pm 1$ (\%) & 99.9 & 99.8  & 99.8 & 100 \\
        \bottomrule
    \end{tabular}
\end{center}
\caption{Abstraction controlling accuracy. C/L: the C-conditional model with L length controlling. Acc $\pm 1$: Accuracy with 1 point/stroke error.}
\label{tab:acc}
\end{table}

\subsection{Evaluation}
\label{eval}

\textbf{Qualitative evaluation}
Our approach achieves class-conditional sketch synthesis, as demonstrated in Figures \ref{fig:banner} and \ref{fig:class}, where our model generates sketches corresponding to given categories. By incorporating abstraction conditioning, we can flexibly control the abstraction level by determining either the length of points or strokes, as shown in Figure \ref{fig:class_abstract}.

As illustrated in Figures \ref{fig:banner}, our photo-to-sketch models demonstrate the ability to capture the semantics of references with various abstraction levels. Overall, our method exhibits the capability of generating high-quality sketches, whether for category-based or instance-based synthesis, over a wide range of abstractions, showcasing its effectiveness and versatility.

\textbf{Quantitative Evaluation}
To demonstrate that our models can accurately control abstraction levels, we provide the accuracy of the length of points and strokes of our models in Table \ref{tab:acc}. As shown in the table, our models exhibit high precision for both class-conditional and photo-to-sketch synthesis, and for both point length and stroke length. It can be observed that almost all the results fall within a slight error, further emphasizing the effectiveness of proposed state embeddings and the stroke token in explicitly controlling sketch length.

\begin{figure}[t]
\begin{center}
    \includegraphics[width=1.\linewidth]{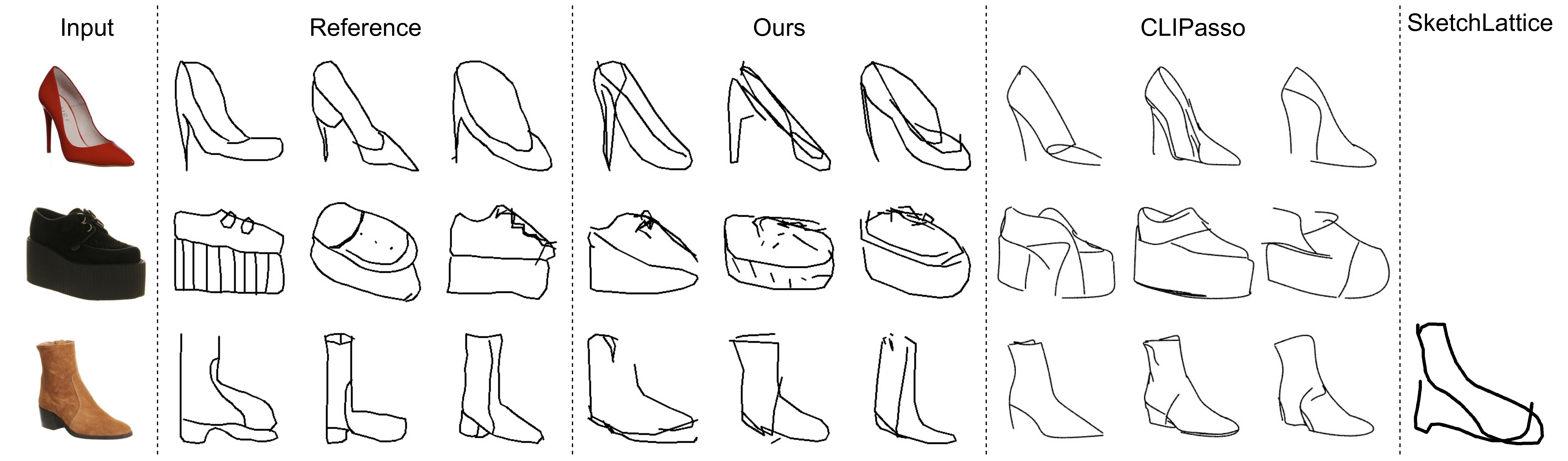}
\end{center}
   \caption{Comparison on photo-to-sketch synthesis. Reference: human sketches in the QMUL Shoe-V2 dataset.}
\label{fig:compare}
\end{figure}

\begin{figure}[t]
\begin{center}
    \includegraphics[width=1.\linewidth]{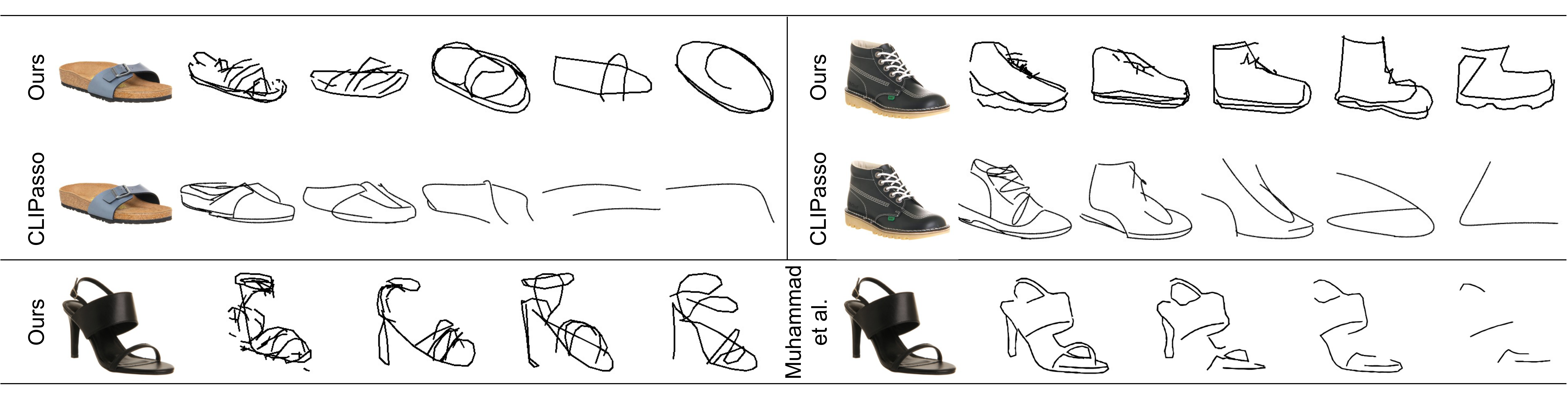}
\end{center}
   \caption{Stroke length controlling comparison. The sketches in the top two rows using 16, 8, 4, 2, and 1 strokes from left to right. We use 32, 16, 8, and 4 strokes in the bottom row.}
\label{fig:compare_stroke}
\end{figure}

\subsection{Qualitative Comparison}
\label{compare_qual}

\textbf{Photo-to-Sketch synthesis}
In Figure \ref{fig:compare}, we compare our results with SketchLattice \cite{sketchlattice} and CLIPasso \cite{clipasso}. Due to the absence of SketchLattice's implementation, we use the results directly from their paper, hence some results are missing. CLIPasso \cite{clipasso} presents results with close geometric shapes to the images; however, the style is significantly different from the reference sketches drawn by humans. This discrepancy arises because their optimization-based algorithm has no opportunity to learn from real sketch vectors. The CLIP-similarity minimization in CLIPasso leads to results that appear monotonous and lack variety. Instead, our model successfully captures the way humans tend to sketch: less precise and ambiguous yet still presenting the semantics of the object. Our results exhibit different styles, as people have various drawing habits. Our model generates diverse and non-rigid sketches, which effectively capture human sketching characteristics.

\textbf{Controlling abstraction}
In Figure \ref{fig:compare_stroke}, we compare our results with two stroke-length controlling works: Muhammad \etal \cite{Muhammad_2018_CVPR} and CLIPasso \cite{clipasso}. Since the implementation of Muhammad \etal is not available, we use the results directly from their paper. Muhammad \etal \cite{Muhammad_2018_CVPR} increase abstraction by removing strokes, which can make the sketches appear incomplete. Additionally, the remaining strokes cannot be adjusted accordingly.

CLIPasso \cite{clipasso} can use different arrangements for different abstraction levels. However, the fixed number of control points in Bézier curves limits the expressiveness of strokes, so when there are too few strokes, CLIPasso cannot express the sketches effectively. In contrast, our model successfully generates recognizable sketches even with only one stroke. In the QMUL Shoe V2 dataset, a sketch has at most 15 strokes. Therefore, when using too many strokes (i.e., out of the training domain), our results will present some noisy strokes, as shown in the leftmost image in the bottom row. Despite this limitation, our method is flexible and suitable for generating sketches with different levels of abstraction while maintaining recognizability.

\begin{table}[t!]
\begin{center}
    \begin{tabular}{lcccc}
        \toprule
        Model  & CLIP score $\uparrow$ & Acc@3 $\uparrow$ & Acc@10 $\uparrow$ & Speed $\downarrow$ \\
        \midrule
        CLIPasso $\#s = 1$  & 45.8        & 0.03         & 0.14         & 90 \\
        CLIPasso $\#s = 4$  & 55.7        & 0.08         & 0.21         & 91 \\
        CLIPasso $\#s = 16$ & 67.9        & 0.31         & 0.61         & 94 \\
        \hdashline
        Our $\#s = 1$       & 46.4 / 52.8 & 0.06 / 0.10  & 0.18 / 0.29  & 18 / 20 \\
        Our $\#s = 4$       & 47.8 / 54.0 & 0.07 / 0.08  & 0.21 / 0.23  & - \\
        Our $\#s = 16$      & 46.6 / 52.5 & 0.06 / 0.10  & 0.15 / 0.26  & - \\
        Our $\#p = 16$      & 43.4 / 47.3 & 0.06 / 0.06  & 0.13 / 0.18  & 17 / 25 \\
        Our $\#p = 32$      & 47.3 / 54.8 & 0.07 / 0.14  & 0.17 / 0.28  & - \\
        Our $\#p = 64$      & 50.1 / 56.7 & 0.10 / 0.13  & 0.25 / 0.38  & - \\
        \hdashline
        Human sketch        & 53.3        & 0.07         & 0.20         &  \\
        \bottomrule
    \end{tabular}
\end{center}
\caption{Quantitative comparison on photo-to-sketch synthesis. For our models, the second figure is the best result from 10 samples with the highest CLIP score for each object. Acc@k: top-k accuracy of FG-SBIR. Speed: inference time in seconds.}
\label{tab:quant_compare}
\end{table}

\subsection{Quantitative Comparison}
\label{compare_quan}

We performed a quantitative comparison between our model and CLIPasso using the CLIP score and zero-shot CLIP fine-grained sketch-based image retrieval (FG-SBIR) as metrics. To ensure a fair comparison, we used the CLIP ViT-L/14 model, which was not employed in either of the methods, to calculate the similarity between generated sketches and corresponding object photos. The evaluation was conducted on 100 objects from the QMUL Shoe-V2 testing dataset. Table \ref{tab:quant_compare} presents the CLIP scores, retrieval accuracies, and inference times for both approaches.

CLIPasso's performance is heavily impacted by the number of strokes, with its CLIP score and retrieval accuracy dropping substantially as fewer strokes are used. Notice that the Bézier curve used in CLIPasso makes the number of strokes directly affect the overall length. On the other hand, our model performs consistently regardless of the number of strokes used. Our point-length controlling model demonstrates a correlation between the number of points used and both the CLIP score and retrieval accuracy, supporting the claim that point count is a more direct influence on abstraction than stroke count.

Moreover, our model's FG-SBIR accuracy is close to the baseline provided by human sketches.
When taking the highest score of 10 samples for each object, our model's performance closely resembles the baseline CLIP score, suggesting its ability to adapt its strategy to more closely align with human sketching patterns across varied points and strokes.

\subsection{Ablation Study}
\label{ablation}

In our approach, we take the point sequence of sketches as input, causing strokes is a more challenging concept than points. This prompts us to conduct an ablation study on stroke length controlling, as shown in Table \ref{tab:stroke_attn}. We compare our stroke token approach with two other mechanisms: (1) summing up with timestep embedding and using adaLN-Zero, and (2) forming a length-one sequence and applying additional cross-attention layers. The cross-attention conditioning has the most extra parameters but performs the worst. In contrast, both stroke token and adaLN-Zero are parameter-efficient. The performance of the proposed stroke token is slightly better in terms of both accuracy and FID \cite{fid}. This highlights the beneficial interaction that MHA provides for the sketch sequence and stroke token for enhanced control. The stroke token not only constrains the synthesis process unilaterally but can also obtain information from the sketch sequence. Beside accurate control, the lower FID indicates that this information exchange, which adaLN-zero can not offer, might be helpful for generating high-quality sketches.

\begin{table}[h]
\begin{center}
    \begin{tabular}{lcccc}
        \toprule
        Type  & Acc. (\%) $\uparrow$ & Acc. $\pm 1$ (\%) $\uparrow$ & FID $\downarrow$ & Extra Params $\downarrow$ \\
        \midrule
        stroke token    & 96.9 & 99.8 & 6.69  &  886K  \\
        adaLN-Zero      & 96.2 & 99.8 & 7.02  &  1.5M \\
        cross-attention & 8.7  & 25.4 & 7.55  &  30M \\
        \bottomrule
    \end{tabular}
\end{center}
\caption{Ablative results on conditioning type of stroke controlling on class-conditional synthesis. The proposed stroke token demonstrates superior control over stroke length while using fewer extra parameters.}
\label{tab:stroke_attn}
\end{table}

\section{Limitation}

Despite demonstrating promising results, our method encounters some limitations. First, we have not thoroughly studied the first-stage VAE, which is functional but restricts our method's overall performance. For example, the current VAE may struggle to capture fine-grained details in more intricate sketches. Developing a more compact and semantically rich sketch latent representation might lead to enhanced outcomes and efficiency with longer, complex sketches. Second, due to a limited dataset, our photo-to-sketch synthesis is constrained to shoes. As collecting free-hand sketches is challenging, exploring alternative learning approaches, such as few-shot meta-learning or leveraging synthetic data, could be a valuable solution to  expand this task to a broader domain.

\section{Conclusion}

In this work, we have for the first time successfully developed a neural network-based method to explicitly control the length of points and strokes in synthesized free-hand sketches, demonstrating diverse natural styles and abstraction degrees akin to human sketches. The introduced LDM framework accomplishes both class-conditional synthesis and photo-to-sketch synthesis. To the best of our knowledge, this is the first instance of a class-conditional sketch synthesis model across a broad range of categories. By providing more flexible control, our approach enhances the real-world applicability of sketch synthesis.

% \bibliography{egbib}

\newpage
\renewcommand\thesection{\Alph{section}}
\setcounter{section}{0}

\begin{center}
    {\Large \bf Supplementary Material}
\end{center}

\section{First Stage VAE}

\begin{figure}[h]
\begin{center}
    \includegraphics[width=1.\linewidth]{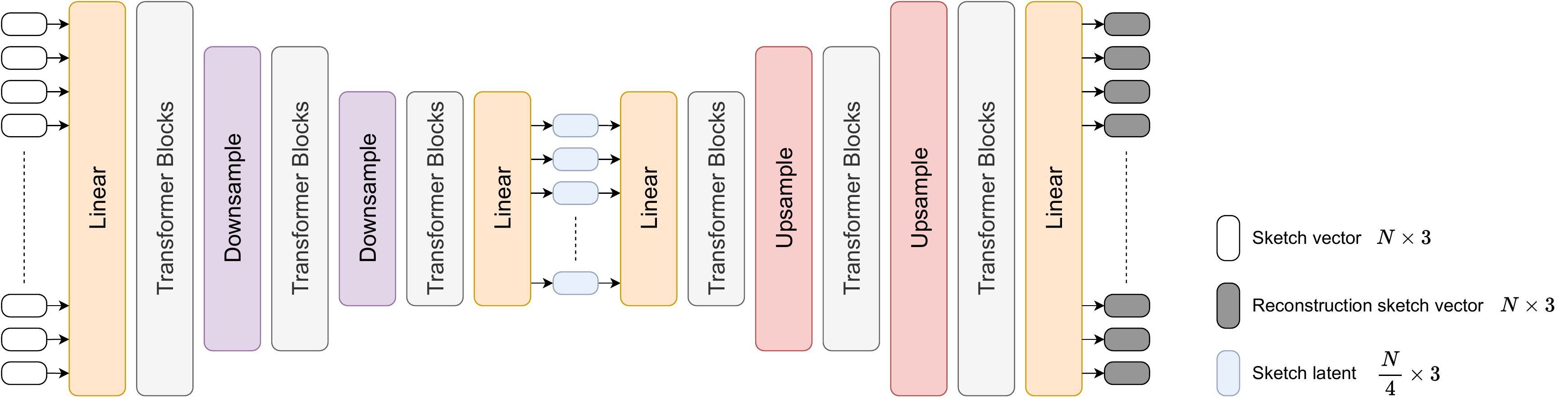}
\end{center}
   \caption{Architecture of first-stage VAE.}
\label{fig:vae_architecture}
\end{figure}

In this section, we demonstrate the architecture and training objective of the first-stage VAE, designed to compress the sketch vector's length dimension.
As Figure \ref{fig:vae_architecture} illustrates, the transformer-based VAE's encoder is composed of a series of transformer blocks interspersed with linear downsampling layers. Mirroring this architecture, the decoder is constructed of a stack of transformer blocks interspersed with linear upsampling layers.
Utilizing a compression ratio of $4$, the sketch latent representation has a reduced dimension of $\frac{N}{4} \times 3$, where the original dimension of the sketch vector is $N \times 3$.

% For the absolute positions $(x_i, y_i)$, we use the $L2$ loss. Binary cross-entropy loss handles the binary pen states $p_i$. To regularize the latent representation $z$, we apply the Kullback-Leibler (KL) loss.
Our losses include $L2$ loss for absolute positions $(x_i, y_i)$, binary cross-entropy loss for binary pen states $p_i$, and Kullback-Leibler (KL) loss for latent $z$ regularization.
Denote $(\hat{x}_i, \hat{y}_i, \hat{p}_i)$ as the reconstruction of $(x_i, y_i, p_i)$ from the VAE.
The combination objective is as follows:

\begin{equation}
    \mathcal{L}_{abs} = \mathbb{E}\left[\lVert (x_i, y_i) - (\hat{x}_i, \hat{y}_i) \rVert_2^2\right]
\end{equation}

\begin{equation}
    \mathcal{L}_{state} = -\mathbb{E}\left[(p_i\log(\hat{p}_i) + (1 - p_i)\log(1 - \hat{p}_i))\right]
\end{equation}

\begin{equation}
    \mathcal{L}_{KL} = \frac{1}{2} \sum (1 + log(\sigma_z^2) - \mu_z^2 - \sigma_z^2)
\end{equation}

\begin{equation}
    \mathcal{L}_{VAE} = \mathcal{L}_{abs} + \mathcal{L}_{state} + \lambda\mathcal{L}_{KL}
\end{equation}

The low-weighted coefficient $\lambda$ is set to $10^{-6}$ in this work.

\section{Additional Qualatative Results}
\begin{figure}[t]
\begin{center}
    \includegraphics[width=1.\linewidth]{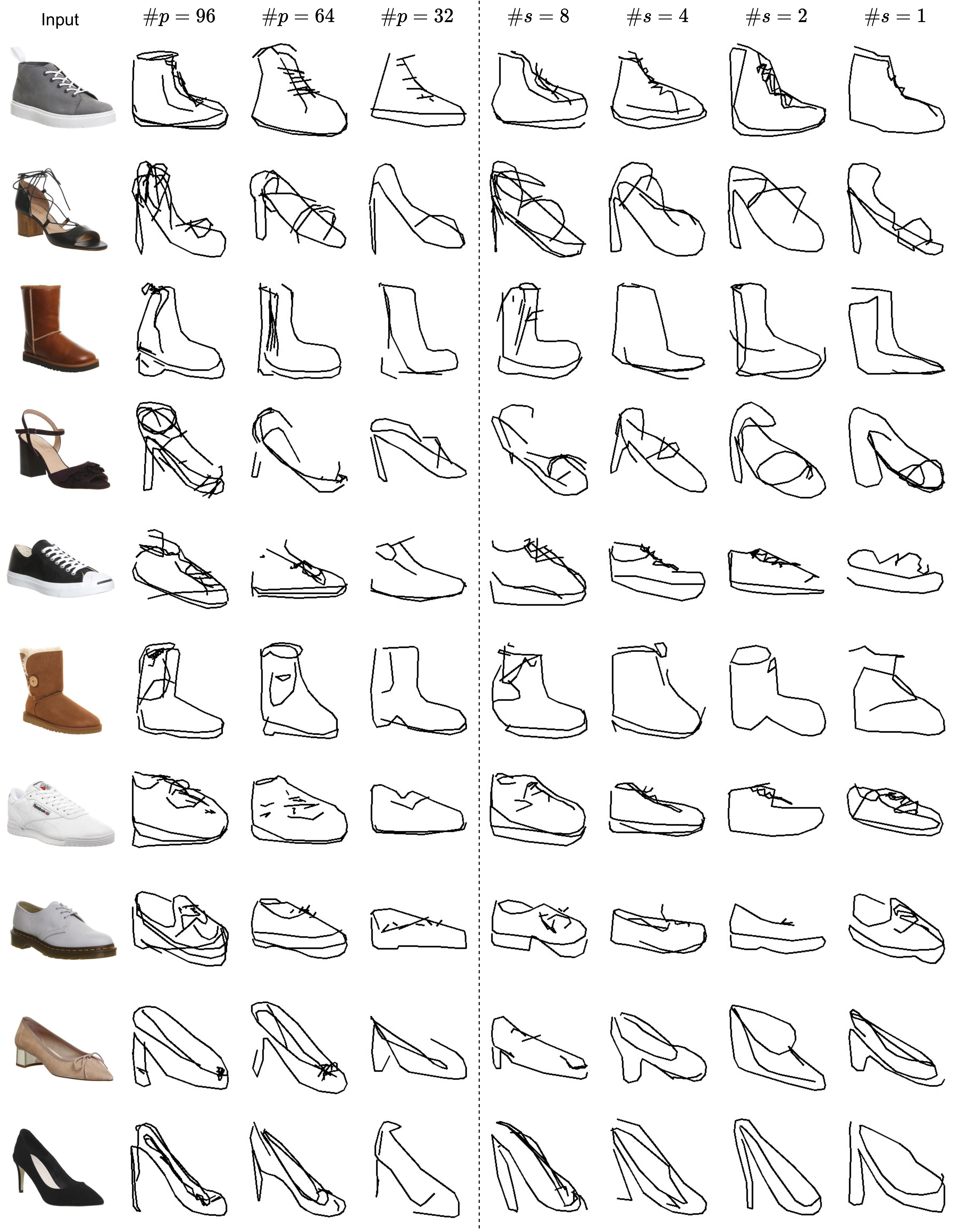}
\end{center}
   \caption{Additional length control results of photo-to-sketch models.}
\label{fig:clip_abstract}
\end{figure}

We provide additional length controlling results of class-conditional models and photo-to-sketch models in Figure \ref{fig:class_abstract} and \ref{fig:clip_abstract}, respectively.

\begin{figure}[t]
\begin{center}
    \includegraphics[width=1.\linewidth]{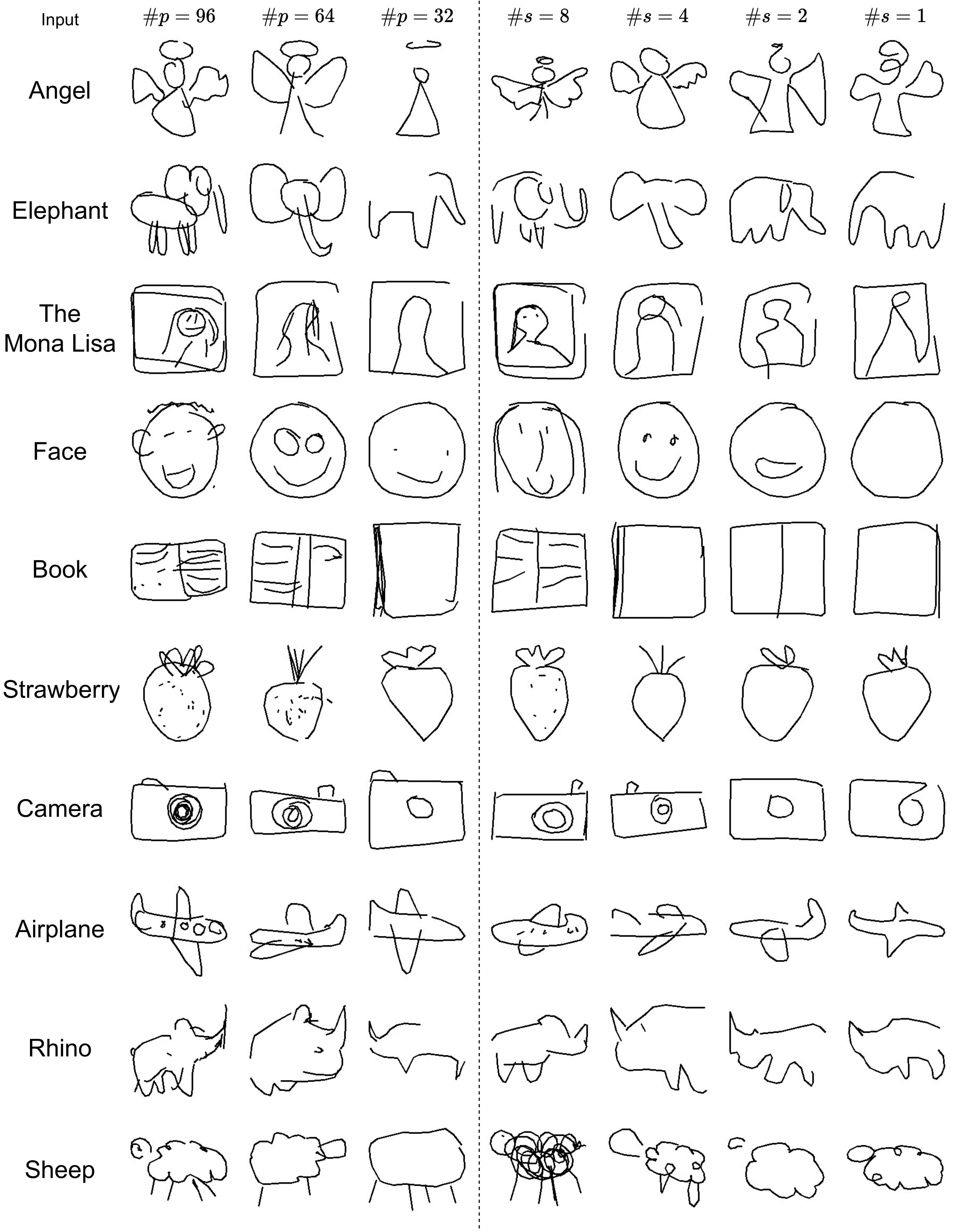}
\end{center}
   \caption{Additional length control results of class-conditional models.}
\label{fig:class_abstract}
\end{figure}

\end{document}